\documentclass[letterpaper]{article} 
\usepackage{aaai24}  
\usepackage{times}  
\usepackage{helvet}  
\usepackage{courier}  
\usepackage[hyphens]{url}  
\usepackage{graphicx} 
\usepackage{natbib}  
\usepackage{caption} 
\usepackage{xcolor}
\usepackage{algorithm}
\usepackage{algorithmic}

\usepackage{newfloat}
\usepackage{listings}
\usepackage{amssymb}
\usepackage{multirow}
\DeclareCaptionStyle{ruled}{labelfont=normalfont,labelsep=colon,strut=off} 
\floatstyle{ruled}
\newfloat{listing}{tb}{lst}{}
\floatname{listing}{Listing}
\title{Synergistic Multiscale Detail Refinement via  \\ Intrinsic Supervision for Underwater Image Enhancement}
\author{
    Dehuan Zhang\textsuperscript{\rm 1}\equalcontrib,
    Jingchun Zhou\textsuperscript{\rm 1}\equalcontrib \thanks{Corresponding author},
    Chunle Guo\textsuperscript{\rm 2},
    Weishi Zhang\textsuperscript{\rm 1},
    Chongyi Li\textsuperscript{\rm 2}
}
\affiliations{
     \textsuperscript{\rm 1} College of Information Science and Technology, Dalian Maritime University\\
    \textsuperscript{\rm 2} VCIP, CS, Nankai University\\
     zhangdehuan97@gmail.com, zhoujingchun03@gmail.com,  guochunle@nankai.edu.cn, 
     teesiv@dlmu.edu.cn, 
     lichongyi@nankai.edu.cn
}

\usepackage{bibentry}

\begin{document}

\maketitle

\begin{abstract}
Visually restoring underwater scenes primarily involves mitigating interference from underwater media.
Existing methods ignore the inherent scale-related characteristics in underwater scenes. Therefore, we present the synergistic multi-scale detail refinement via intrinsic supervision (SMDR-IS) for enhancing underwater scene details, which contain multi-stages. The low-degradation stage from the original images furnishes the original stage with multi-scale details, achieved through feature propagation using the Adaptive Selective Intrinsic Supervised Feature (ASISF) module. By using intrinsic supervision, the ASISF module can precisely control and guide feature transmission across multi-degradation stages, enhancing multi-scale detail refinement and minimizing the interference from irrelevant information in the low-degradation stage. In multi-degradation encoder-decoder framework of SMDR-IS, we introduce the Bifocal Intrinsic-Context Attention Module (BICA). Based on the intrinsic supervision principles, BICA efficiently exploits multi-scale scene information in images. BICA directs higher-resolution spaces by tapping into the insights of lower-resolution ones, underscoring the pivotal role of spatial contextual relationships in underwater image restoration. Throughout training, the inclusion of a multi-degradation loss function can enhance the network, allowing it to adeptly extract information across diverse scales. When benchmarked against state-of-the-art methods, SMDR-IS consistently showcases superior performance. The code is publicly available at: \url{https://github.com/zhoujingchun03/SMDR-IS}
\end{abstract}

\section{Introduction}
In the complex dynamics of underwater environments, the quality of optical images is primarily determined by the influence of dissolved and suspended substances on light absorption and scattering \cite{URanker}. Absorption effects lead to challenges like reduced imaging distance and color distortion, while scattering effects diminish image contrast and detail. Our goal is to enhance the quality of underwater optical images, providing robust solutions for applications, such as underwater exploration, marine biology research, and surveillance \cite{liu4}. Image enhancement technique empowers researchers and practitioners to more effectively interpret and analyze underwater image data \cite{jiangqiuping2}.

Enhancing low-quality images poses significant challenges to the field of computer vision \cite{add5,add4,IACC}. These challenges arise primarily from scattering and blurring effects unique to aquatic environments, which inherently manifest in a multi-scale manner \cite{zhouIJCV}. Particulate matter and water turbulence at different scales have different effects on different scales have different effects on different parts of an image, leading to the degradation of multi-scale correlated features. The underwater image formation model (UIFM) \cite{joe} is represented as:
\begin{equation}
I = J \times t + A(1 - t)
\end{equation}
where $I$ represents the clear image, $J$ is the underwater image, $t$ represents the transmission related to depth, and $A$ denotes the atmospheric light. 

\begin{figure}[!t]
\centering
\includegraphics[width=0.8\columnwidth]{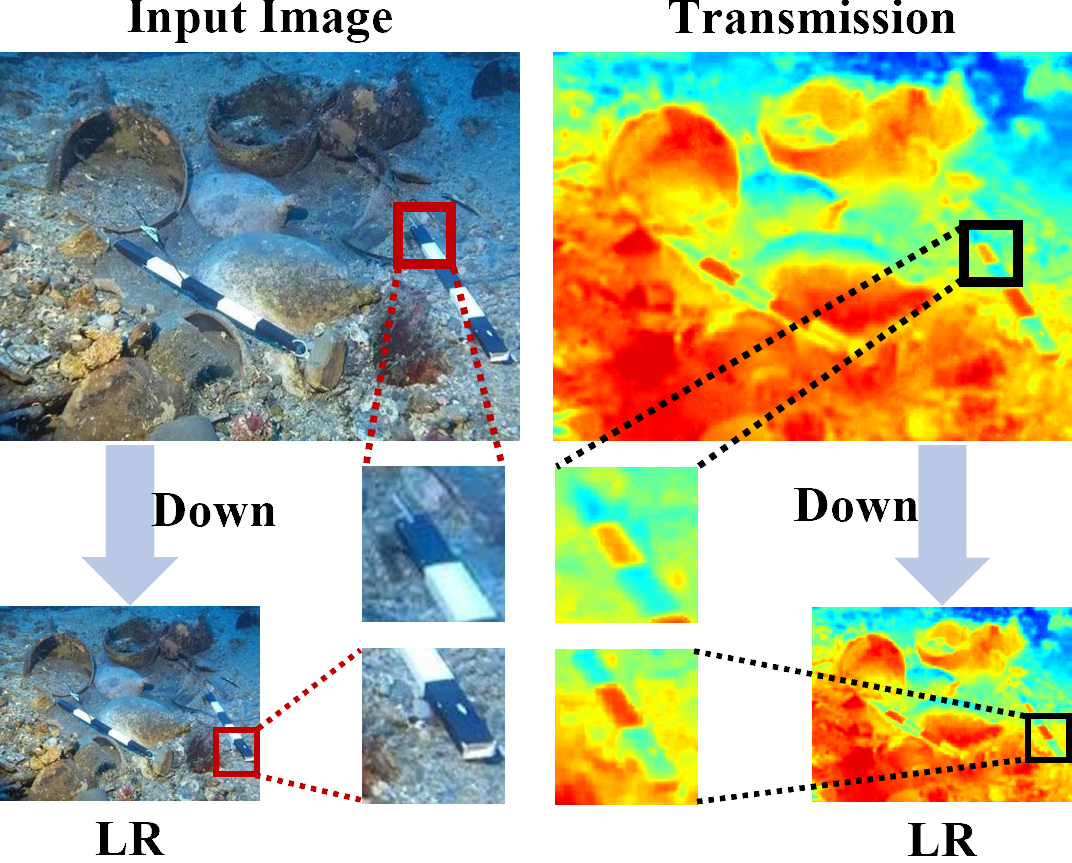} 
\caption{Illustration of motivation. The figure showcases similar scene information extracted from multi-resolution images alongside transmission. Degradation patterns, consistent across different positions, are evident in both the original and the downscaled images.}
\label{figure_mulit}
\end{figure}

The core objective of underwater image enhancement (UIE) is to accurately estimate $I$ from Eq. (1) \cite{zhou1}. This is crucial because degradation levels in underwater images can differ across pixels due to variations in distance. There are many existing enhancement techniques \cite{zhuang2}, which either globally or locally extract features \cite{liu1} \cite{liu2}. These methods often leverage filters \cite{zhou2} and color correction techniques \cite{zhou_colorbalance} to improve image quality.  
As illustrated in Figure \ref{figure_mulit},  similar degradation scenarios exist in different resolution stages, Therefore, the inherent scale-related characteristics in multi-scale underwater scenes \cite{jiangkui}\cite{hou1} demand attention. 
By recognizing inherent scale-related degradation patterns, we can gain a richer understanding of the structure of the scene. In light of this, we proposed Synergistic Multiscale Detail Refinement via Intrinsic Supervision for Underwater Image Enhancement (SMDR-IS). SMDR-IS uniquely leverages low-resolution images as auxiliary inputs, providing additional insights into scene degradation.

SMDR-IS simultaneously harnesses multiple degradations to mitigate the loss of multiple scene features resulting from feature extractions. While feature extraction at the original yields high-level features produce high-level features, it is often possible to capture finer data features at the expense of larger-scale scene information. The low-degradation stages serve to counteract the excessive abstraction from redundant extractions, ensuring that the intrinsic relationship between features and the original image remains intact. By integrating low-degradation images and features from the initial layers, both credibility and applicability are enhanced. Furthermore, the incorporation of ASISF guarantees that irrelevant information is excluded from the original resolution.

The contributions are summarized as follows:

(1) Addressing the limitations of scale-related features in current underwater image enhancement methods, which lead to incomplete restoration of scene details, we propose a synergistic multiscale detail refinement via intrinsic supervision for underwater image enhancement.

(2) We design a new attention, namely Bifocal Intrinsic-Context Attention, based on a dual-path approach, ensuring that both feature enhancement and contextual semantic information are addressed. Additionally, we integrate resolution-guided supervision to boost computational efficiency without sacrificing detail enhancement quality.

(3) With the aim of refining low-resolution feature information, we introduce the Adaptive Selective Intrinsic Supervised Feature (ASISF) module. ASISF regulates feature propagation, enhancing image quality and avoiding blurring caused by utilizing the overlay of multiscale scene details.

(4) The integration of a multi-degradation loss function provides constraints and optimization for learning features at each stage. This approach empowers the network to effectively exploit information at various scales, thereby improving detail and structure recovery.

\section{Related Work}

UIE techniques have become increasingly prevalent in the domains of computer vision and image processing \cite{zhuang1}. Broadly speaking, deep learning-based UIE methods can be divided into prior-based methods and end-to-end methods.

\subsection{Prior-based Image Enhancement}
Prior-based methods rely on explicit degradation models or pre-existing knowledge to calibrate model parameters. For instance, UColor \cite{Ucolor} which is guided by transmission, employs a transmission-driven image enhancement network using GDCP \cite{GDCP}. \cite{sunjiaming} uses an augmented U-Net to fuse inputs, such as the original, color-corrected, and contrast-enhanced images, to effectively leverage features. Unsupervised underwater image restoration method (USUIR) \cite{USUIR} involves designing a transmission subnet, a global background subnet, and a scene radiance subnet to estimate parameters of UIFM, facilitating Photo-realistic image restoration. Zhang et al. \cite{Rex-net} proposed Rex-Net, which applied the Retinex theory to enhance underwater images. 
\cite{add3} employs physical knowledge to design the Adaptive Transmission-Guided Module to guide the network.
While these methods have shown promising results, the intricate and unpredictable nature of underwater environments sometimes undermines the efficacy of the proposed priors, potentially limiting model adaptability.

\subsection{End-to-end Image Enhancement}
End-to-end UIE techniques circumvent the need for explicit degradation models or prior knowledge. UIEC$^2$ \cite{UIEC} enhanced image in the RGB and HSV color spaces. UIEC$^2$ employs RGB pixel-level blocks for color restoration and adjusts saturation using curves in the HSV domain. In \cite{zhou_multiview}, an enhancement method is devised based on cross-view images, which employs feature alignment to fuse scene information from multiple perspectives. 
Liu et al. \cite{liu3} put forth an image-aware adversarial fusion network rooted in object detection. This strategy integrates a multi-scale dense enhancement subnet to bolster visual results. 
\cite{add4} proposed a lightweight transformer method (UIEPTA), which presents a gray-scale attention model to guide the network for extracting non-contaminated features.
Notwithstanding their merits, many existing end-to-end methods tend to overlook the full potential of scale-related features in image scenes. This oversight often results in difficulties in precise scene detail recovery. To address this gap, we introduce SMDR-IS, emphasizing synergistic multiscale detail refinement and intrinsic supervision.

\section{Methodology}
\begin{figure*}
\centering
\includegraphics[width=1.8\columnwidth]{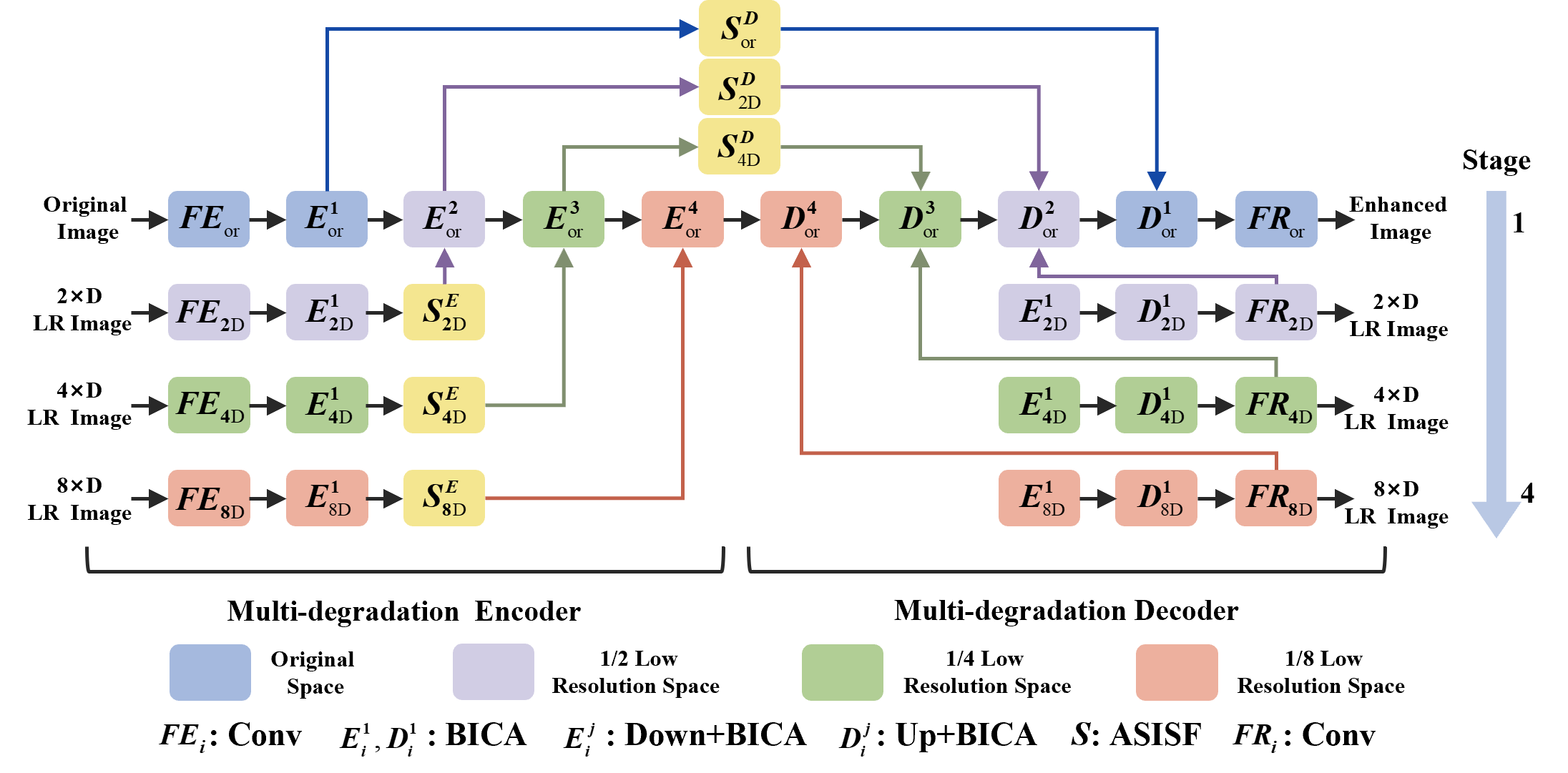} 
\caption{The overview of SMDR-IS. In Decoder, $E_i^1$ is the same as in Encoder.
$FE_i$, $i \in \{ or,2D,4D,8D \} $ is Feature Extraction (Conv). $E_i^1$, $D_i^1$ denote the Bifocal Intrinsic-Context Attention (BICA). $E_i^j$, $j \in [2,4]$ is Downsamling+BICA. $D_i^j$ is Upsamling+BICA. $S$ denotes Adaptive Selective Intrinsic Supervised Feature Module (ASISF). $FR_i$ is Feature Restoration (Conv).}
\label{network}
\end{figure*}

We propose Synergistic Multiscale Detail Refinement via Intrinsic Supervision for Underwater Image Enhancement (SMDR-IS), as depicted in Figure \ref{network}. SMDR-IS comprises a multi-degradation encoder and decoder. 

\subsection{Multi-Degradation Encoder}
Our proposed model integrates four stages of multi-resolution image inputs, equipping the network and the related features with diverse scale scene information of the input image. Initially, we employ downsampling on the original image to derive underwater scenes of different scales. 
Since prevailing UNet-based underwater enhancement methods \cite{Ucolor} typically utilize three downsampling steps, we adopt three downscaled stages corresponding to each stage of the UNet. This can capture a richer array of scale scene information. To harness the scale-correlated attributes of image scenes, SMDR-IS proposed the multi-degradation encoder, further supplemented by three stages for lower-resolution images.

Each encoder stage incorporates the Bifocal Intrinsic-Context Attention Module (BICA) to extract feature details. BICA can adeptly fuse global and local features, optimizing image enhancement. With the aid of resolution-guided supervision, BICA strikes a harmonious balance between computational efficiency and details enhancement.
To bolster the utility of low-resolution features, we proposed the Adaptive Selective Intrinsic Supervised Feature (ASISF) module. This module governs feature propagation, elevating image enhancement outcomes, and curbing information blurring that may arise from superimposing multi-scale scene details.

\begin{figure*}[!ht]
\centering
\includegraphics[width=1.8\columnwidth]{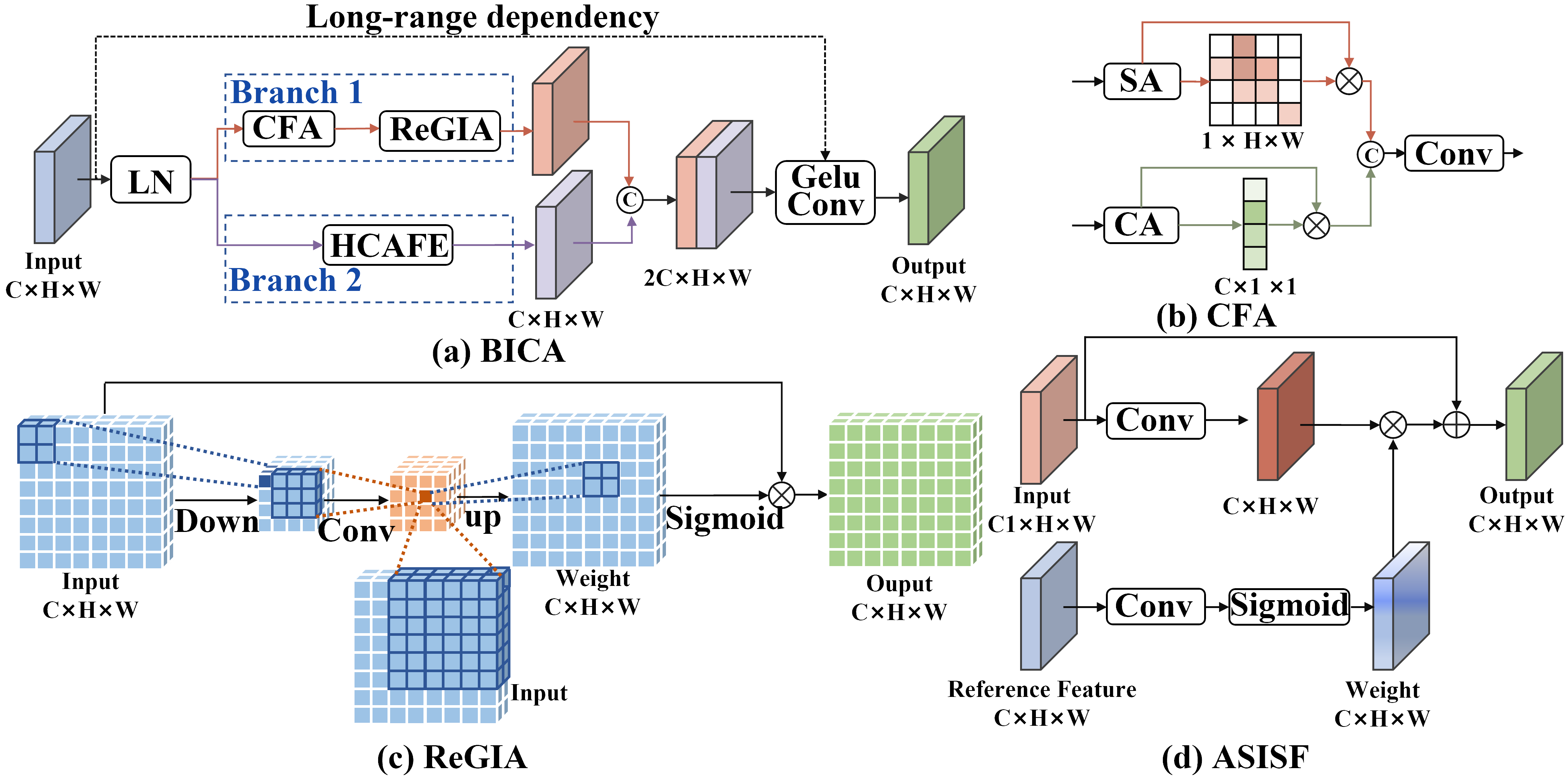} 
\caption{(a) Architecture of BICA, LN represents Layer Normalization. (b) CFA, CA and SA represent Channel Attention and Spatial Attention, respectively. (c) ReGIA, "Down" and "Up" denote downsampling and upsampling, respectively. In low-resolution, $1\times1$ corresponds to $6\times6$ spatial feature in the original resolution. (d) ASISF, highlighting that the channels in both input and reference features can be diverse.}
\label{BICA}
\end{figure*}

\noindent 
\textbf{Bifocal Intrinsic-Context Attention} 
The bifocal intrinsic-context attention module (BICA), as depicted in Figure \ref{BICA}, consists of two distinct branches. The first branch is responsible for recognizing the significant influence of neighboring pixel regions for image restoration. This branch contains the Comprehensive Feature Attention (CFA) module (detailed in the supplementary material), followed by the Resolution-Guided Intrinsic Attention (ReGIA) module. The CFA module, as shown in Figure \ref{BICA} (b), extracts features from pixels and channels through spatial attention and channel attention, respectively. The features from CFA are fed to the ReGIA module, whose function is to broaden the receptive field while maintaining computational efficiency. As shown in Figure \ref{BICA}(c), by using low-resolution spatial intrinsic supervision, ReGIA can further enrich the features by effectively capturing multi-scale scene details.

In the second branch, acknowledging the pivotal role of spatial contextual relationships in underwater image restoration, we design the Hierarchical Context-Aware Feature Extraction (HCAFE) module (detailed in the supplementary material). This module functions within the original feature domain, extracting image representation features through hierarchical context attention.

BCIA leverages a resolution-guided supervision approach, achieving superior computational efficiency without sacrificing detail quality. BCIA employs an innovative dual-path attention mechanism, comprehensively addressing both the influence of neighboring pixel regions and spatial contextual relationships. 


\noindent
\textbf{Resolution-Guided Intrinsic Attention Module}
Underwater image optimization is heavily influenced by the contextual details of adjacent pixels. However, to reduce the computational efficiency of the network, enhancement methods often utilize compact $3\times3$ convolutional kernels for feature extraction. Although $3\times3$ kernels are computationally efficient, the small receptive field hinders the network to capture extensive contextual features.

To address the limitation and broaden the receptive field, we propose the Resolution-Guided Intrinsic Attention Module (ReGIA), which enhances the preliminary features extracted from the input image by CFA. ReGIA is tailored to discern feature weight data in a lower-resolution latent space with a large feature receptive field. Guided by the lower-resolution latent space, ReGIA serves as a guiding beacon that not only enhances the correlation of features in the original domain but also strives to improve the computational efficiency of the network.

\noindent
\textbf{Adaptive Selective Intrinsic Supervised Feature Module} 
When integrating feature information from the low-resolution encoder and decoder into the original resolution branch, it is imperative to mitigate interference from non-essential information during image restoration \cite{MPNet}. To this end, we introduce the Adaptive Selective Intrinsic Supervised Feature Module (ASISF). This module adopts an intrinsic-supervised approach for feature selection and constraint, as depicted in Figure \ref{BICA}.

ASISF is meticulously designed to retain only the most pertinent features to supplement image feature extraction and reconstruction in the original resolution stage. When learning features from the low-resolution encoder, the features from the original resolution act as a reference. 
This strategy ensures that both the encoder and decoder components of multi-degradation align with the genuine degradation traits inherent in underwater images.

This is particularly significant in the context of image restoration, where the presence of noise, distortions, and varying levels of degradation can complicate the restoration process. By employing intrinsic supervision, ASISF effectively identifies and retains features meaningful for the restoration task, thereby improving the precision and efficiency of the restoration process. The feature adaptive selection is pivotal for effective learning and generalization of the SMDR-IS, ultimately enhancing image restoration capabilities by supplementing inherent scale-related degradation patterns at multi-resolution.

\subsection{Multi-Degradation Decoder}
Unlike the traditional UNet, the multi-degradation decoder in SMDR-IS is uniquely designed to complement its encoder counterpart. Low-degradation information captures the inherent scale-related characteristics of input images. By integrating this information, the decoder can gain a deeper insight into these degradation patterns and multiple scenes.

The fusion of low-degradation and residual information from the encoder equips the decoder with a holistic grasp of the complicated image degradation. This comprehensive understanding empowers the network to produce restoration results that are both precise and context-rich. Essentially, the low-degradation information acts as the supplementary branch, enhancing the decoder's restoration capabilities. 
Consequently, the output images not only exhibit crisp details but also accurately recover information from multiple scenes. It is worth mentioning that within the $FR_{iD}$, the ASISF performs intrinsic-supervised feature selection on the output features when feeding data to the original degradation encoder. This ensures that only the most relevant feature is propagated through the image processing pipeline. In essence, the multi-degradation decoder of SMDR-IS uses low-degradation cues to ingeniously supplement inherent scale-related characteristics, thereby notably improving the fidelity of the restored images.

\begin{figure*}[!ht]
\centering
\includegraphics[width=2.1\columnwidth]{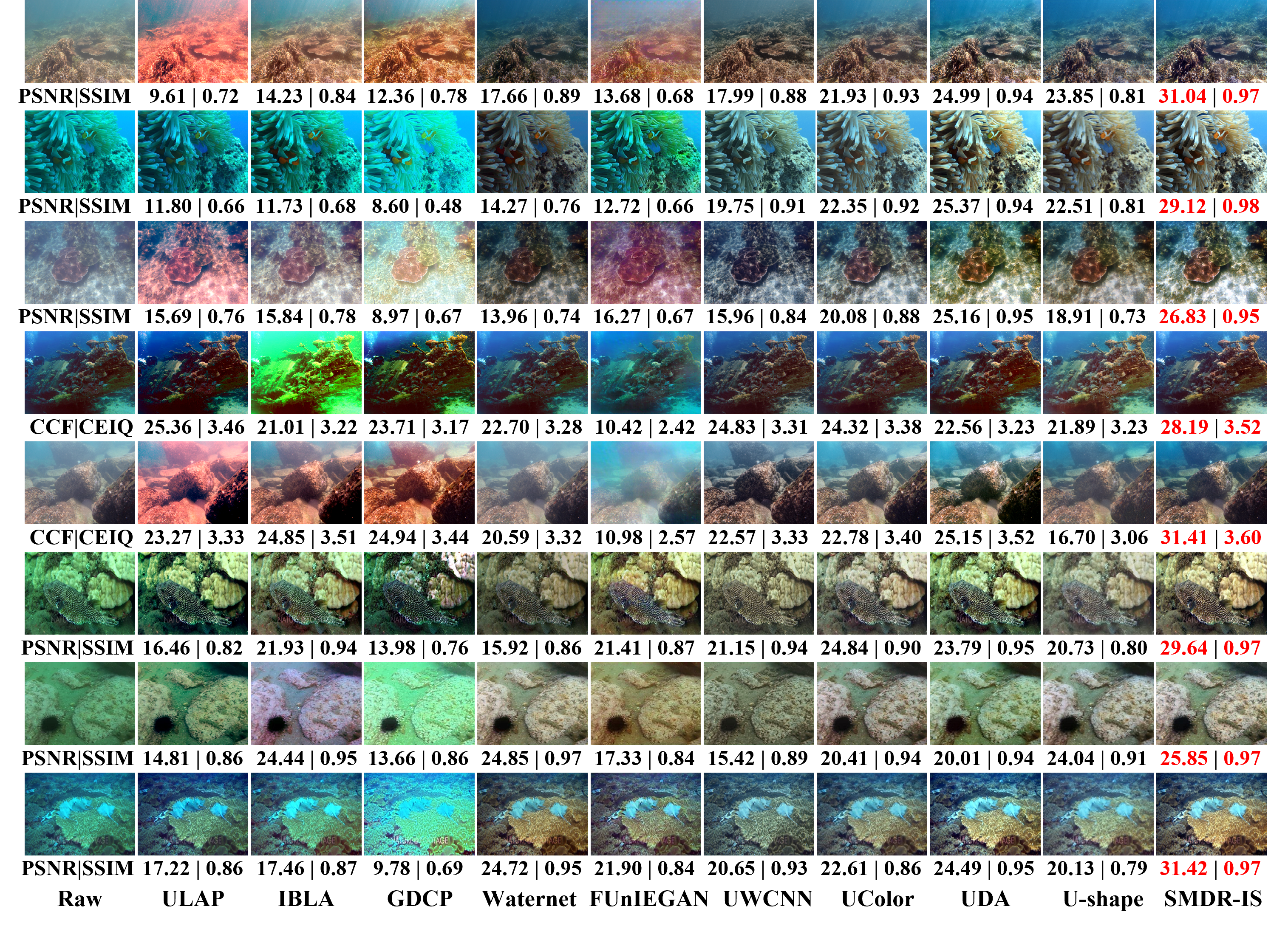} 
\caption{Qualitative comparison between state-of-the-art methods and the SMDR-IS on various datasets.}
\label{fig_experiment}
\end{figure*}

\subsection{Multi-Degradation Loss Function}
To enhance the precision of the restoration process, SMDR-IS employs a multi-degradation loss function to govern different modules of the network, which compares the multi-resolution enhancement image of SMDR-IS with multi-resolution ground truth, to instill multi-resolution supervision. 

\begin{equation}
L_1 = \sum_{i=1}^{N} |y_i - \hat{y}_i|
\end{equation}

\begin{equation}
L_{pre} =  \| \phi(y)_i - \phi(\hat{y})_i \|_2^2
\end{equation}

\begin{equation}
L_{mse} = \frac{1}{N} \sum_{i=1}^{N} (y_i - \hat{y}_i)^2
\end{equation}

\begin{equation}
L_{j} =  L_1^j+ 0.2 \times L_{pre}^j+L_{mse}^j
\end{equation}
where $N$ is the total number of pixels in the image, $y_i$ denotes the ground-truth pixel value, $\hat{y}_i$ represents the predicted pixel value. $\phi$ is the function in VGGNet. $i$ represents the $i$-th layer's feature maps in the pre-trained network. $L_{j}$ represents the loss function of the \textit{j}-th stage, $j$ is an index, from 1 to 4. 4 is the number of stages in SMDR-IS. 

The overall loss functions ($L$) for training SMDR-IS can be written as follows:
\begin{equation}
L = \sum_{j=1}^{4} L_{j}
\end{equation}

\begin{table*}[htbp]
\centering
    \fontsize{9pt}{10pt}\selectfont
    \begin{tabular}{cccccccccccc}
\hline
        Dataset & Metric & ULAP & IBLA & GDCP & WaterNet & 
        {\begin{tabular}[c]{@{}c@{}}FUnIE\\ \begin{tabular}[c]{@{}c@{}}GAN\\ \end{tabular} \end{tabular}}
          & UWCNN  & UColor  & UDA & U-shape & SMDR-IS \\ \hline
       \multirow{11}{*}{\begin{tabular}[c]{@{}c@{}}UIEB\\ \begin{tabular}[c]{@{}c@{}}Val\\ \end{tabular} \end{tabular}}  & PSNR $\uparrow$ & 15.913 & 17.988 & 13.386 & 17.349 & 17.114 & 17.985 & 20.962  & 21.484 & 20.462 &  \textbf{23.710}  \\ 
         & MSE $\downarrow$ & 0.174 & 0.143 & 0.228 & 0.144 & 0.150 & 0.134 & 0.097  & 0.096 & 0.100 & \textbf{0.075}  \\ 
         & SSIM $\uparrow$& 0.745 & 0.805 & 0.747 & 0.813 & 0.701 & 0.844 & 0.863  & 0.873 & 0.792 & \textbf{0.922}  \\ 
         & VSI $\uparrow$& 0.947 & 0.958 & 0.943 & 0.966 & 0.941 & 0.966 & 0.971  &  0.970  & 0.959 &\textbf{0.983}  \\ \
         & FSIM $\uparrow$ & 0.915 & 0.928 & 0.901 & 0.908 & 0.891 & 0.923 & 0.931  & 0.932 & 0.892 & \textbf{0.967}  \\ 
         & FSIMc $\uparrow$ & 0.878 & 0.899 & 0.865 & 0.899 & 0.858 & 0.903 & 0.920   & 0.918 & 0.883 & \textbf{0.957}  \\ 
         & UIQM $\uparrow$& 2.259 & 2.490 & 2.670 & 2.917 & 3.092 & 3.011 & 3.049   & 2.897 & 3.131 & \textbf{3.015}  \\ 
         & UCIQE $\uparrow$ & 0.604 & 0.606 & 0.592 & 0.606 & 0.564 & 0.554 & 0.591  & \textbf{0.612} & 0.576 & 0.607  \\ 
         & CCF $\uparrow$ & 24.145 & 23.841 & 23.026 & 20.042 & 20.416 & 20.360 & 21.827  & \textbf{26.220} & 21.966 & 26.012  \\ 
         & CEIQ $\uparrow$ & 3.209 & 3.283 & 3.208 & 3.101 & 3.307 & 3.090 & 3.209  & \textbf{3.372} & 3.235 & 3.369  \\ \cline{2-12}
         & ALL $\uparrow$ & 49.441 & 51.656 & 46.109 & 47.455 & 47.733 & 48.502 & 53.226  & 58.374 & 52.997 & \textbf{60.466 } \\ \hline
         \multirow{5}{*}{U45}
         & UIQM $\uparrow$& 2.282 & 2.388 & 2.275 & 2.957 & 2.495 & 3.064 & 3.148  & 2.878 & 3.175	 & \textbf{3.121}  \\ 
         & UCIQE $\uparrow$& 0.588 & 0.595 & 0.597 & 0.601 & 0.545 & 0.554 & 0.586   &\textbf{0.607} & 0.571 & 0.605 \\ 
         & CCF $\uparrow$& 22.069 & 21.598 & 22.736 & 20.391 & 12.931 & 21.418 & 22.100  &  25.449 & 21.284  & \textbf{25.489}  \\ 
         & CEIQ $\uparrow$& 3.192 & 3.249 & 3.191 & 3.186 & 2.785 & 3.213 & 3.283  & 3.392 & 3.254 & \textbf{3.397}  \\ \cline{2-12}
         & ALL $\uparrow$& 28.131 & 27.830 & 28.799 & 27.135 & 18.757 & 28.248 & 29.117  &  32.326 & 28.285 & \textbf{32.612}  \\ \hline
 \multirow{11}{*}{LSUI}  & PSNR $\uparrow$  &  17.677 & 17.555 & 13.284 & 19.990 & 21.129 & 20.368 & 21.786  & 20.288 & 20.491 & \textbf{21.984}  \\ 
         & MSE $\downarrow$ & 0.142 & 0.149 & 0.229 & 0.109 & 0.101 & 0.102 & 0.087  & 0.104 & 0.102 & \textbf{0.089}   \\ 
         & SSIM $\uparrow$ & 0.760 & 0.784 & 0.710 & 0.839 & 0.778 & 0.851 & 0.848  & 0.842 & 0.775 & \textbf{0.870}   \\ 
         & VSI $\uparrow$  & 0.959 & 0.957 & 0.935 & 0.971 & 0.962 & 0.970 & 0.973  & 0.968 & 0.960 & \textbf{0.975}  \\ 
         & FSIM $\uparrow$ & 0.925 & 0.925 & 0.886 & 0.931 & 0.920 & 0.940 & 0.940  & 0.931 & 0.900 & \textbf{0.948}    \\ 
         & FSIMc $\uparrow$ & 0.894 & 0.894 & 0.850 & 0.918 & 0.900 & 0.921 & 0.928  & 0.912 & 0.888 & \textbf{0.933}   \\ 
         & UIQM $\uparrow$ & 2.324 & 2.567 & 2.572 & 2.865 & 2.895 & 2.976 & 2.984 & 2.833 & \textbf{3.038} & 2.917   \\ 
         & UCIQE $\uparrow$    & 0.611 & 0.616 & 0.610 & 0.605 & 0.585 & 0.561 & 0.593 & \textbf{0.617} & 0.576 & 0.602   \\ 
         & CCF $\uparrow$  & 24.004 & 24.135 & 24.304 & 20.531 & 21.484 & 21.254 & 22.331 & \textbf{25.942} & 21.849 & 24.727   \\ 
         & CEIQ $\uparrow$  & 3.160 & 3.282 & 3.227 & 3.123 & 3.195 & 3.147 & 3.259 & \textbf{3.361} & 3.246 & 3.323   \\ \cline{2-12}
         & ALL $\uparrow$   & 51.457 & 51.864 & 47.606 & 50.882 & 52.949 & 52.091 & 54.729 & 56.799 & 52.826 & \textbf{57.368}  \\ \hline
    \end{tabular}
\caption{Quantitative comparison between state-of-the-art methods and SMDR-IS on different testing datasets.}
\label{table_all_expreiment}
\end{table*}

\section{Experiments}\label{Experiments}
\subsection{Datasets} \label{Dataset}

We trained SMDR-IS utilizing the UIEB dataset, which consists of 800 training images and 90 paired testing images. To assess the robustness of SMDR-IS, we further conducted the evaluation on various datasets, i.e. UIEB, U45, LSUI.

\subsection{Implementation Details} 
Our method was implemented using PyTorch, with an NVIDIA Tesla V100 GPU, Intel(R) Xeon(R) Silver 4114 CPU, and 32GB RAM. For training, images were randomly cropped to a resolution of 256$\times$256, and we used a batch size of 44 and a learning rate of 0.0002. To ensure that SMDR-IS can generate outputs consistent with the original image size during testing, we adopted the border padding techniques.

\subsection{Comparison Results}
In this section, we adopted both objective assessments (UIQM, UCIQE \cite{jiangqiuping}, CCF \cite{ccf}, CEIQ \cite{ceiq}, VSI \citep{VSI}, PSNR \citep{PSNR}, MSE, SSIM \citep{SSIM}, FSIM, FSIMc \citep{FSIM}) and subjective evaluations for comprehensive analysis. 

We conducted experiments to compare SMDR-IS with state-of-the-art methods to underscore its effectiveness. This includes traditional methods, such as ULAP \cite{ULAP}, IBLA \cite{IBLA}, GDCP \cite{GDCP}, as well as deep learning methods, such as WaterNet \cite{WaterNetUieb}, FUnIEGAN \cite{FastGan}, UWCNN \cite{UWCNN}, UColor \cite{Ucolor}, UDA \cite{UDAformer} and U-shape \cite{ushape}. The visual results are illustrated in Figure \ref{fig_experiment}, and the metrics are shown in Table \ref{table_all_expreiment}.

To avoid any influence of color on the metrics (e.g., the impact observed with ULAP in Figure \ref{fig_experiment}), we excluded the color component from CCF. As shown in Figure \ref{fig_experiment}, SMDR-IS performs better than other methods in terms of visual results and performance metrics. Although ULAP, IBLA, and GDCP achieve commendable restoration results, their generalization capabilities are hindered by their dependence on priors. WaterNet and UColor display enhanced robustness but remain susceptible to extraneous inputs. Although FUnIE-GAN and UWCNN can achieve real-time efficiency, their expressive capability is somewhat constrained due to limited parameters.
Although UDA and U-shape can both achieve the goal of underwater image enhancement, SMDR-IS combines the inherent scale-related features from multiple scales, providing better robustness for diverse underwater scenarios.

Table \ref{table_all_expreiment} tabulates the performance of different methods on the three datasets, in terms of different performance metrics. In addition to the individual performance metrics, we also combine them to form a comprehensive metric, denoted as "ALL", which is the net sum of the $\uparrow$ values minus the $\downarrow$ values. This measurement can provide a holistic assessment by weighing multiple criteria. Remarkably, SMDR-IS achieves the best performance, in terms of the ALL score. Collectively, these results show the ability of SMDR-IS to adeptly restore intricate scenes, underpinned by the synergy from multiscale detail extraction and intrinsic supervision.

\begin{table*}[htp]
\centering
\fontsize{9pt}{11pt}\selectfont
    \begin{tabular}{ccccccccccc}
\hline
          & ULAP & IBLA & GDCP & WaterNet &  
          FUnIEGAN
          & UWCNN  & UColor   & UDA &  U-shape &  SMDR-IS \\ \hline
      Time $\downarrow$ & 0.358 &  9.134 & 0.163 & 0.091 & \textbf{0.003} &  0.050 &  0.577  & 0.100 & 0.109 & 0.061  \\ 
       Quality $\uparrow$ &49.441 & 51.656 & 46.109 & 47.455 & 47.733 & 48.502 & 53.226  & 58.374 & 52.997 & \textbf{60.466}  \\ \hline
       Agg $\uparrow$ & 49.083 &42.522&45.946&47.364&47.730&	48.452&52.649 & 58.273 &  52.887 & \textbf{60.405} \\ \hline
    \end{tabular}
\caption{Aggregative evaluation of efficiency and image quality.}
\label{table_time}
\end{table*}

\begin{table}[!ht]
\centering
    \begin{tabular}{ccc|ccc}
    \hline
        En & En\_to\_DE & De & PSNR $\uparrow$ & SSIM $\uparrow$ & ALL $\uparrow$  \\ \hline
        ~ & \checkmark   & \checkmark   & 23.021 & 0.915 & 23.936 \\ 
        \checkmark   & ~ & \checkmark   & 23.697  & 0.919  & 24.615  \\ 
        \checkmark   & \checkmark   &    & 23.122  & 0.914  & 24.036  \\ 
        \checkmark   & \checkmark   & \checkmark   & \textbf{23.710}  & \textbf{0.922}  & \textbf{24.631}  \\ \hline 
    \end{tabular}
            \caption{Ablation study for ASISF.}
\label{table_ASISF}
\end{table}
\begin{table}[!ht]
\centering
    \begin{tabular}{cccc|ccc}
    \hline
        S1 & S2 & S3 & S4 & PSNR $\uparrow$ & SSIM $\uparrow$ & ALL $\uparrow$ \\ \hline
        \checkmark   & ~ & ~ & ~ & 22.790 & 0.916 & 23.706  \\ 
        \checkmark   & \checkmark   & ~ & ~ & 22.872 & 0.915  & 23.787  \\ 
        \checkmark   & \checkmark   & \checkmark   & ~ & 23.248 & 0.915  & 24.163  \\ 
        \checkmark   & \checkmark   & \checkmark   & \checkmark   & \textbf{23.710}  & \textbf{0.922}  & \textbf{24.631}  \\ \hline 
    \end{tabular}
\caption{Ablation study for the four different stages. S represents the stage.}
\label{table_stage}
\end{table}

\begin{table}[!ht]
\centering
\resizebox{\columnwidth}{!}{
    \begin{tabular}{cccc|ccc}
    \hline
        CA & SA & ReGIA & HCAFE & PSNR $\uparrow$ & SSIM $\uparrow$ & ALL $\uparrow$ \\ \hline
        ~ & \checkmark   & \checkmark   & \checkmark   & 21.219  & 0.901  & 22.120  \\ 
        \checkmark   & ~ & \checkmark   & \checkmark   & 23.352  & 0.917  & 24.269  \\ 
        \checkmark   & \checkmark   & ~ & \checkmark   & 22.673  & 0.907  & 23.580  \\ 
        \checkmark   & \checkmark   &  \checkmark & ~ & 23.276  & 0.911  & 24.186  \\ 
        \checkmark   & \checkmark   & \checkmark   & \checkmark  & \textbf{23.710}  & \textbf{0.922}  & \textbf{24.631}  \\ \hline 
    \end{tabular}
    }
\caption{Ablation study for BICA.}
\label{table_BICA}
\end{table}

\begin{table}[!ht]
\centering
     \begin{tabular}{ccc|ccc}
    \hline
        $L_1$ & $L_{pre}$ & $L_{mse}$ & PSNR $\uparrow$ & SSIM $\uparrow$ & ALL$\uparrow$ \\ \hline
        ~ & \checkmark   & \checkmark   & 23.257  & 0.915  & 24.172  \\ 
        \checkmark   & ~ & \checkmark   & 22.905 & 0.912  & 23.817 \\ 
        \checkmark   & \checkmark   & ~ & 23.151 & 0.919 & 24.069 \\ 
        \checkmark   & \checkmark   & \checkmark   & \textbf{23.710}  & \textbf{0.922}  & \textbf{24.631}  \\ \hline 
    \end{tabular}
\caption{Ablation study for loss function. }
\label{table_loss}
\end{table}

\begin{figure}[!ht]
\centering
\includegraphics[width=1\columnwidth]{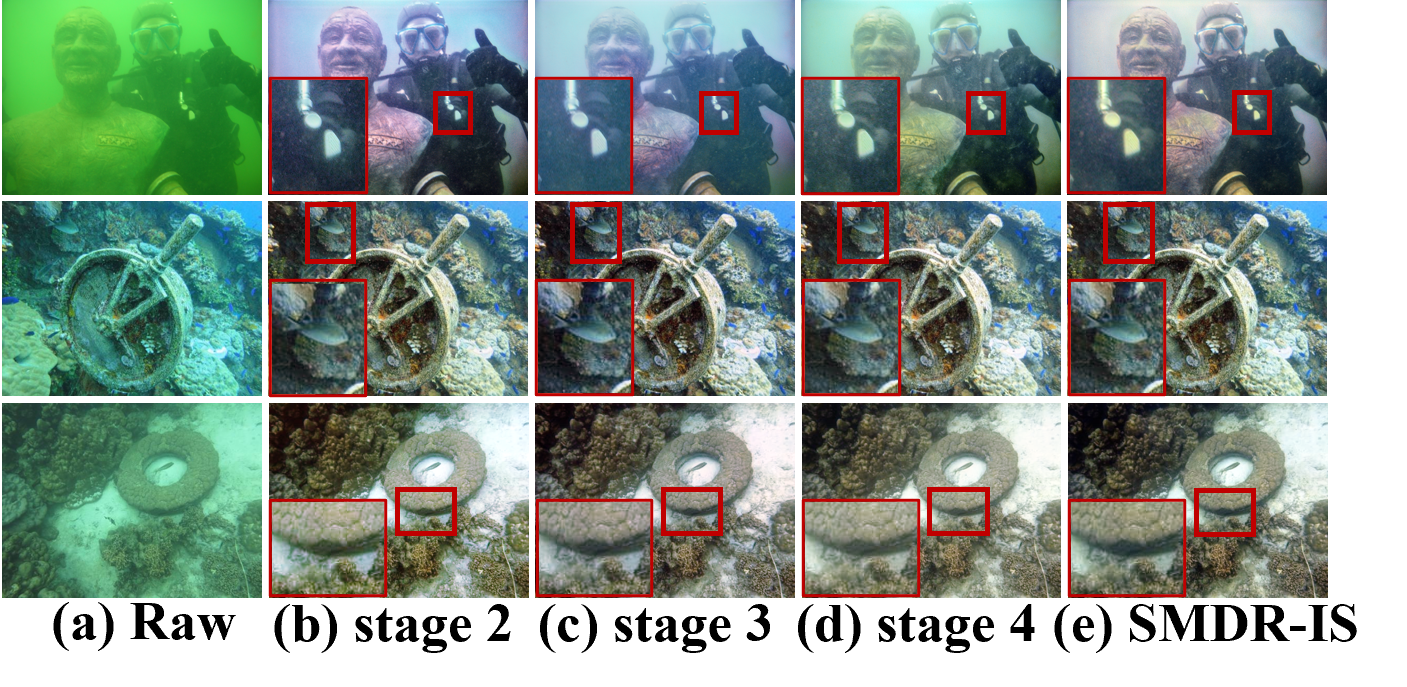} 
\caption{Subjective results of ablation experiments on number of stages used. From left to right: (a) original images, (b)-(e) correspond to lines 1-4 in Table \ref{table_stage}.}
\label{stage_ablation}
\end{figure}

\subsection{Efficiency Evaluation}
Table \ref{table_time} presents the comprehensive analysis of efficiency, performance (the Quality is the "ALL" on UIEB Val in Table \ref{table_all_expreiment}), and Aggregative (Agg) score. Although SMDR-IS does not obtain the highest efficiency, it still achieves a commendable frame rate of 16.4744, satisfying real-time demands. Moreover, we introduce the 'Agg' score to provide a holistic performance evaluation, which considers both efficiency and performance, specifically, Agg = Quality-Time. As observed from Table \ref{table_time}, SMDR-IS achieves the highest score in Agg, which demonstrates the suitability of SMDR-IS for advanced underwater vision tasks.

\subsection{Ablation Study}\label{Ablation Study}

We performed a series of ablation experiments using the testing set from the UIEB dataset. The effects of degradation at various stages are presented in Table \ref{table_stage}. Furthermore, we conducted ablation studies on the individual components within BICA, as detailed in \ref{table_BICA}, on the ASISF component within SMDI-IR, as illustrated in Table \ref{table_ASISF}, and on the different loss functions, as showcased in Table \ref{table_loss}. In all tables, entries in bold represent the highest scores obtained. From Table \ref{table_ASISF}, it is evident that the intrinsic guidance provided by ASISF plays a pivotal role in filtering out irrelevant features during the image enhancement process.

As illustrated in Table \ref{table_stage} and Figure \ref{stage_ablation}, the number of stages has a significant impact on the performance metrics. When increasing the number of stages, the image restoration performance and the multi-scale feature extraction capability of SMDI-IR progressively improve. The performance metrics reach the peak when the number of stages is four. It is worth noting that our choice of four stages is informed by prevalent practices, especially three-fold downsampling is commonly utilized, as highlighted in \cite{Ucolor}.

The advantages of the BICA architecture are manifestly demonstrated in Table \ref{table_BICA}, where the ablation experiments highlight the image restoration ability of each module within BICA. Notably, the contribution of the proposed ReGIA to SMDI-IR is particularly significant, reaffirming the superior capabilities of ReGIA. Furthermore, to ascertain the efficacy of the loss functions used in our study, we conducted ablation experiments by sequentially omitting $L_1$, $L_{pre}$, and $L_{mse}$ in Table \ref{table_loss}. These experiments collectively substantiate the effectiveness of our chosen loss functions.

\section{Conclusion}
In this study, we propose a novel method for underwater image restoration, namely SMDR-IS, which can proficiently capture multi-scale scene information, by multi-resolution detail extraction with intrinsic supervision, and utilizing the inherent scale-related features in image scenes. To ensure optimal assimilation of information across different scales, we integrated low-resolution inputs by adaptive selective intrinsic supervision within the original-resolution input, thereby amplifying the conveyance of scene information. To mitigate unnecessary interference from irrelevant scenes, we introduced ASISF, which is meticulously designed to regulate the feature propagation process.
Furthermore, our multi-degradation loss function strategically guides the network during training. Optimization and constraints at each stage enhance the network's ability to leverage information at different scales. 
In the future, we will explore the integration of SMSR-IS into broader computer vision applications, like underwater robotics and autonomous underwater vehicles.

\section{Acknowledgments}

This work was supported in part by the Natural Science Foundation of China (Nos. 62301105), National Key Research and Development Program of China (No.2018AAA0100400), China Postdoctoral Science Foundation (NO.2021M701780), and the Cultivation Program for the Excellent Doctoral Dissertation of Dalian Maritime University. We are also sponsored by CAAI-Huawei MindSpore Open Fund and the High Performance Computing Center of Dalian Maritime University.

\bigskip

\bibliography{AAAI}

\end{document}